\documentclass[11pt,a4paper]{article}
\usepackage{etoolbox}
\usepackage[hyperref]{emnlp2020}
\usepackage{times}
\usepackage{latexsym}

\usepackage{graphicx}
\usepackage{paralist}
\usepackage{bbm}
\usepackage{pifont}
\usepackage{xspace}
\usepackage{xcolor,colortbl}
\usepackage{comment}
\usepackage{amsmath}
\usepackage{svg}
\usepackage{textcomp,gensymb}

\usepackage{microtype}

\aclfinalcopy 
\def\aclpaperid{647} 

\title{Where Are You? Localization from Embodied Dialog}

\author{
Meera Hahn$^1$\thanks{\enspace Correspondence: meerahahn@gatech.edu}\quad
Jacob Krantz$^2$ \quad
Dhruv Batra$^{1,3}$ \quad
Devi Parikh$^{1,3}$ \\
\textbf{James M. Rehg$^1$ \quad
Stefan Lee$^2$ \quad
Peter Anderson$^1$\thanks{\enspace Now at Google.}}\\
[0.1in]
$^1$Georgia Institute of Technology \quad
$^2$Oregon State University \quad
$^3$Facebook AI Research (FAIR)\\
[0.15in]
{\small Project Webpage:} \url{https://meerahahn.github.io/way/}
}
\usepackage{graphicx} 
\usepackage{booktabs} 
\usepackage{xparse}   
\NewDocumentCommand{\rot}{O{-45} O{1em} m}{\makebox[#2][l]{\hspace{-3em}\rotatebox{#1}{#3}}}%

\begin{document}

\newcommand{\dataset}{\textsc{Where Are You?}\xspace}
\newcommand{\acronym}{\textsc{WAY}\xspace}
\newcommand{\Obs}{Observer\xspace}
\newcommand{\Loc}{Locator\xspace}
\newcommand{\obs}{observer\xspace}
\newcommand{\loc}{locator\xspace}

\newcommand{\figref}[1]{Fig.~\ref{#1}}
\newcommand{\tabref}[1]{Tab.~\ref{#1}}
\newcommand{\algref}[1]{Alg.~\ref{#1}}
\newcommand{\secref}[1]{Sec.~\ref{#1}}
\newcommand{\myqoute}[1]{\emph{`#1'}}

\newcommand{\SUPP}[1]{\textul{SUPP: #1}}

\newcommand{\xhdr}[1]{\vspace{3pt}\noindent\textbf{#1}}
\newcolumntype{s}{>{\columncolor[gray]{.85}[.1\tabcolsep]}c}

\maketitle
\begin{abstract}
We present \dataset (\acronym), a dataset of $\sim$6k dialogs in which two humans -- an Observer and a Locator -- complete a cooperative localization task. The Observer is spawned at random in a 3D environment and can navigate from first-person views while answering questions from the Locator. The Locator must localize the Observer in a detailed top-down map by asking questions and giving instructions. Based on this dataset, we define three challenging tasks: Localization from Embodied Dialog or LED (localizing the Observer from dialog history), Embodied Visual Dialog (modeling the Observer), and Cooperative Localization (modeling both agents). In this paper, we focus on the LED task -- providing a strong baseline model with detailed ablations characterizing both dataset biases and the importance of various modeling choices. Our best model achieves 32.7\% success at identifying the Observer's location within 3m in unseen buildings, vs. 70.4\% for human Locators.
\end{abstract}

\section{Introduction}
Imagine getting lost in a new building while trying to visit a friend who lives or works there. Unsure of exactly where you are, you call your friend and start describing your surroundings (\myqoute{I'm standing near a big blue couch in what looks like a lounge. There are a set of wooden double doors opposite the entrance.}) and navigating in response to their questions (\myqoute{If you go through those doors, are you in a hallway with a workout room to the right?}). After a few rounds of dialog, your friend who is familiar with the building will hopefully know your location. Success at this cooperative task requires goal-driven questioning based on your friend's understanding of the environment, unambiguous answers communicating observations via language, and active perception and navigation to investigate the environment and seek out discriminative observations.

\begin{figure}[t]
\centering
\includegraphics[width=1 \columnwidth]{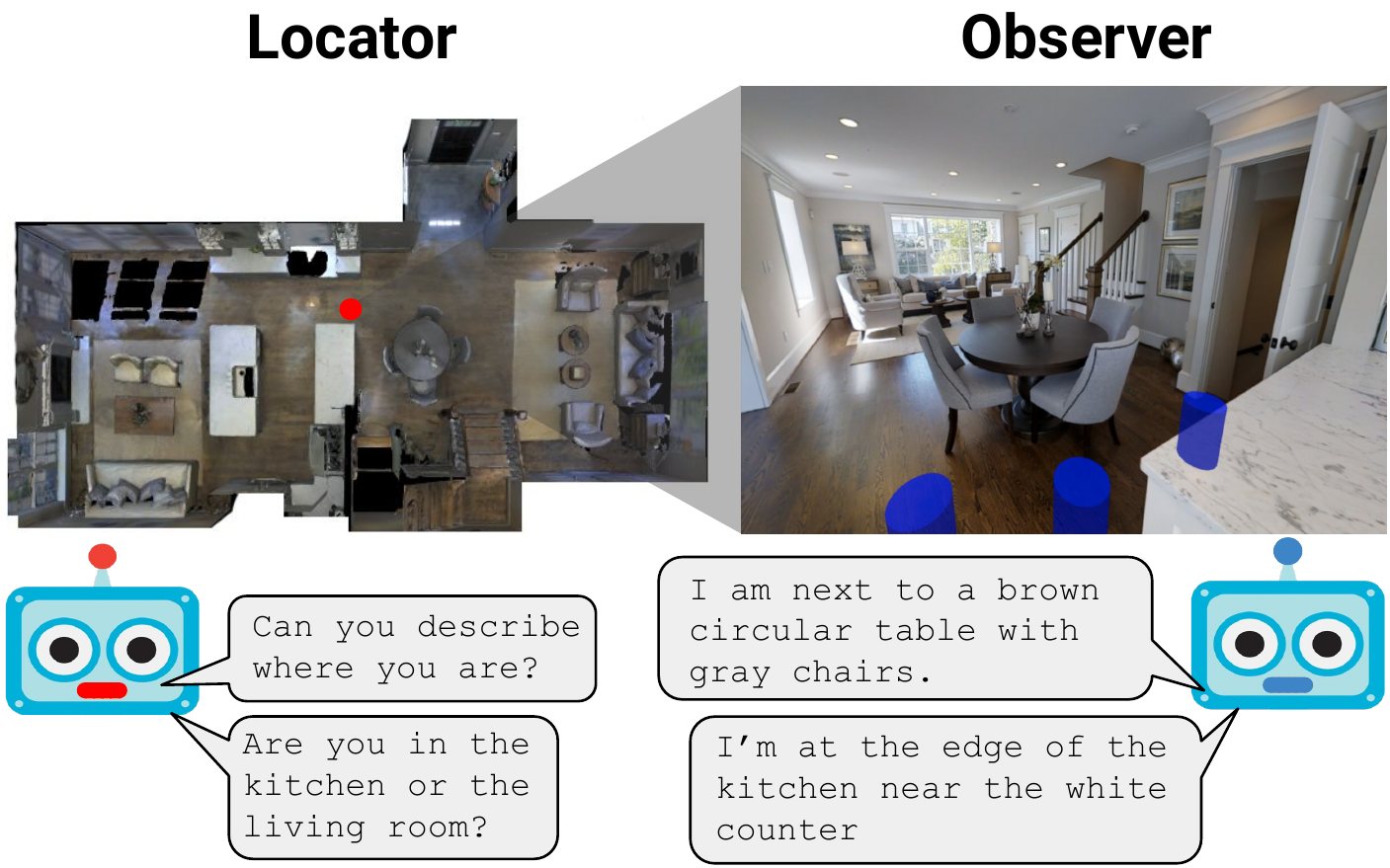}
\caption{LED Task: The Locator has a top-down map of the building and is trying to localize the Observer by asking questions and giving instructions. The Observer has a first person view and may navigate while responding to the Locator. The turn-taking dialog ends when the Locator predicts the Observer's  position.}
\label{fig:task_teaser}
\end{figure}

In this work we present \dataset (\acronym), a new dataset based on this scenario. As shown in Fig. \ref{fig:task_teaser}, during data collection we pair two annotators: an Observer who is spawned at random in a novel environment, and a Locator who must precisely localize the Observer in a provided top-down map. The map can be seen as a proxy for familiarity with the environment -- it is highly detailed, often including multiple floors, but does not show the Observer's current or initial location. In contrast to the ``remote'' Locator, the Observer navigates within the environment from a first-person view but without access to the map. To resolve this information asymmetry and complete the task, the Observer and the Locator communicate in a live two-person chat. The task concludes when the Locator makes a prediction about the current location of the Observer. For the environments we use the Matterport3D dataset \cite{chang2017matterport3d} of 90 reconstructed indoor environments. In total, we collect $\sim$6K English dialogs of humans completing this task from over 2K unique starting locations.

The combination of localization, navigation, and dialog in \acronym provides for a variety of modeling possibilities. We identify three compelling tasks encapsulating significant research challenges: 

\xhdr{-- Localization from Embodied Dialog.} LED, which is the main focus of this paper, is the state estimation problem of localizing the Observer given a map and a partial or complete dialog between the Locator and the Observer. Although localization from dialog has not been widely studied, we note that indoor localization plays a critical role during calls to emergency services \cite{falcon2018predicting}. As 3D models and detailed maps of indoor spaces become increasingly available through indoor scanners \cite{chang2017matterport3d}, LED models could have the potential to help emergency responders localize emergency callers more quickly by identifying locations in a building that match the caller's description. 

\xhdr{-- Embodied Visual Dialog.} EVD is the navigation and language generation task of fulfilling the Observer role. This involves using actions and language to respond to questions such as \myqoute{If you walk out of the bedroom is there a kitchen on your left?}
In future work we hope to encourage the transfer of existing image-based conversational agents \cite{das2017visual} to more complex 3D environments additionally requiring navigation and active vision, in a step closer to physical robotics. The \acronym dataset provides a testbed for this. 

\xhdr{-- Cooperative Localization.} In the CL task, both the Observer and the Locator are modeled agents. Recent position papers \cite{baldridge2018points, mcclelland2019extending, biskarxiv20} have called for a closer connection between language models and the physical world. However, most reinforcement learning for dialog systems is still text-based \cite{li2016deep} or restricted to static images \cite{das2017learning, de2017guesswhat}. Here, we provide a dataset to warm-start and evaluate goal-driven dialog in a realistic embodied setting.

Our main modeling contribution is a strong baseline model for the LED task based on LingUnet \cite{misra2018mapping}. In previously unseen test environments, our model successfully predicts the Locator's location within 3 meters 32.7\% of the time, vs.~70.4\% for the human Locators using the same map input, with random chance accuracy at 6.6\%. We include detailed studies highlighting the importance of data augmentation and residual connections. Additionally, we characterize the biases of the dataset via unimodal (dialog-only, map-only) baselines and experiments with shuffled and ablated dialog inputs, finding limited potential for models to exploit unimodal priors. 

\noindent\textbf{Contributions}: To summarize:
\begin{compactenum}
    \item We present \acronym, a dataset of $\sim$6k dialogs in which two humans with asymmetric information complete a cooperative localization task in reconstructed 3D buildings. 
    \item We define three challenging tasks: Localization from Embodied Dialog (LED), Embodied Visual Dialog, and Cooperative Localization. 
    \item Focusing on LED, we present a strong baseline model with detailed ablations characterizing both modeling choices and dataset biases.
\end{compactenum}

\section{Related Work}
\noindent\textbf{Image-based Dialog} Several datasets grounding goal-oriented dialog in natural images have been proposed. The most similar settings to ours are Cooperative Visual Dialog \cite{das2017visual,das2017learning}, in which a question agent (Q-bot) attempts to guess which image from a provided set the answer agent (A-bot) is looking at, and GuessWhat?! \cite{de2017guesswhat}, in which the state estimation problem is to locate an unknown object in the image. Our dataset extends these settings to a situated 3D environment allowing for active perception and navigation on behalf of the A-bot (Observer), and offering a whole-building state space for the Q-bot (Locator) to reason about. 

\xhdr{Embodied Language Tasks.} A number of `Embodied AI' tasks combining language, visual perception, and navigation in realistic 3D environments have recently gained prominence, including Interactive and Embodied Question Answering ~\cite{das2018embodied,gordon2018iqa}, Vision-and-Language Navigation or VLN \cite{anderson2018vision, chen2019touchdown, mehta2020retouchdown, qi2019reverie}, and challenges based on household tasks \cite{puig2018virtualhome,ALFRED20}. While these tasks utilize only a single question or instruction input, several papers have extended the VLN task -- in which an agent must follow natural language instructions to traverse a path in the environment -- to dialog settings. \citet{nguyen2019help} consider a scenario in which the agent can query an oracle for help while completing the navigation task. However, the closest work to ours is Cooperative Vision-and-Dialog Navigation (CVDN) \cite{thomason2019vision}. CVDN is a dataset of dialogs in which a human assistant with access to visual observations from an oracle planner helps another human complete a navigation task. CVDN dialogs are set in the same Matterport3D buildings \cite{chang2017matterport3d} and like ours they are goal-oriented and easily evaluated. The main difference is that we focus on localization rather than navigation. Qualitatively, this encourages more descriptive utterances from the first-person agent (rather than eliciting short questions). Our work is also related to Talk the Walk \cite{de2018talk} which presented a dataset for a similar task in an outdoor setting using a restricted, highly-abstracted map which encouraged language that is grounded in the semantics of building types rather than visual descriptions of the environment. 

Table \ref{table:dataset_comparison} compares the language in WAY against existing embodied perception datasets. Specifically, size, length and the density of different parts of speech (POS) are shown. Vocab size was determined by the total number of unique words. We used the \cite{loper2002nltk} POS tagger to calculate the POS densities over the text in each dataset. We find that WAY has a higher density of adjectives, nouns, and prepositions than related datasets suggesting the dialog is more descriptive than the text in existing datasets.

\setlength{\tabcolsep}{5pt}
\begin{table*}[t]
\begin{center}
\caption{Comparison of the language between the WAY dataset and related embodied perception datasets.}\vspace{-0.125in}
\footnotesize
\resizebox{\textwidth}{!}{
\label{table:dataset_comparison}
\begin{tabular}{l cccccccc}
\toprule
Method         & & \scriptsize Dataset Size   & 
\scriptsize Vocab Size   & 
\scriptsize Avg Text Length &
\scriptsize Noun Density &
\scriptsize Adj Density &
\scriptsize Preposition Density &
\scriptsize Dialog
\\ \toprule
CVDN     && 2050 & 2165 &  52 &  0.20 &  0.06  &  0.09 &  Yes\\
TtW     && 10K & 7846 & 110 & 0.20 &  0.07 & 0.11 & Yes \\
VLN     && 21K & 3459 & 29 & 0.27 & 0.03 & 0.17 & No \\\midrule
WAY  && 6154 & 5193 & 61 & 0.30 & 0.12 &  0.18 &  Yes \\ \bottomrule
\end{tabular}}
\end{center}
\end{table*}

\xhdr{Localization from Language.} While localization from dialog has not been intensively studied, localization from language has been studied as a sub-component of instruction-following navigation agents \cite{blukis2018mapping,anderson2019chasing,Blukis:19drone-sureal}. The LingUnet model -- a generic language-conditioned image-to-image network we use as the basis of our LED model in Section \ref{sec:model} -- was first proposed in the context of predicting visual goals in images \cite{misra2018mapping}. This also illustrates the somewhat close connection between grounding language to a map and grounding referring expressions to an image \cite{kazemzadeh2014referitgame,mao2016generation}. 

It is important to note that localization is often a precursor to navigation -- one which has not been addressed in existing work in language-based navigation. In both VLN and CVDN, the instructions are conditioned on specific start locations -- assuming the speaker knows the navigator's location prior to giving directions. The localization tasks of the WAY dataset fill this gap by introducing a dialog-based means to localize the navigator. This requires capabilities such as describing a scene, answering questions, and reasoning about how discriminative potential statements will be to the other agent.

\section{\dataset Dataset}
\label{sec:dataset}

We present the \dataset (\acronym) dataset consisting of 6,134 human embodied localization dialogs across 87 unique indoor environments. 

\xhdr{Environments.}
We build \acronym on Matterport3D \cite{chang2017matterport3d}, which contains 90 buildings captured in 10,800 panoramic images. Each building is also provided as a reconstructed 3D textured mesh. This dataset provides high-fidelity visual environments in diverse settings including offices, homes, and museums -- offering numerous objects to reference in localization dialogs. We use the Matterport3D simulator \cite{anderson2018vision} to enable first-person navigation between panoramas. 

\xhdr{Task.} 
A \acronym episode is defined by a starting location (i.e.~a panorama $p$) in an environment $e$. The \Obs is spawned at $p_0$ in $e$ and the \Loc is provided a top-down map of $e$ (see \figref{fig:task_teaser}). Starting with the \Loc, the two engage in a turn-based dialog ($L_0,O_0, \dots L_{T-1}, O_{T-1}$) where each can pass one message per turn. The \Obs may move around in the environment during their turn, resulting in a trajectory ($p_0, p_1, \dots, p_T$) over the dialog. The \Loc is not embodied and does not move but can look at the different floors of the house at multiple angles. The dialog continues until the \Loc uses their turn to make a prediction ($\hat{p}_T$) of the \Obs's current location ($p_T$). The episode is successful if the prediction is within $k$ meters of the true \emph{final} position -- i.e.~$||p_T - \hat{p}_T||_2 < k$ m. This does not depend on the  initial position, encouraging movement to easily-discriminable locations.

\begin{figure*}[t]
\centering
\includegraphics[clip=true, trim=0in 0.1in 0in 0in,width=1\textwidth]{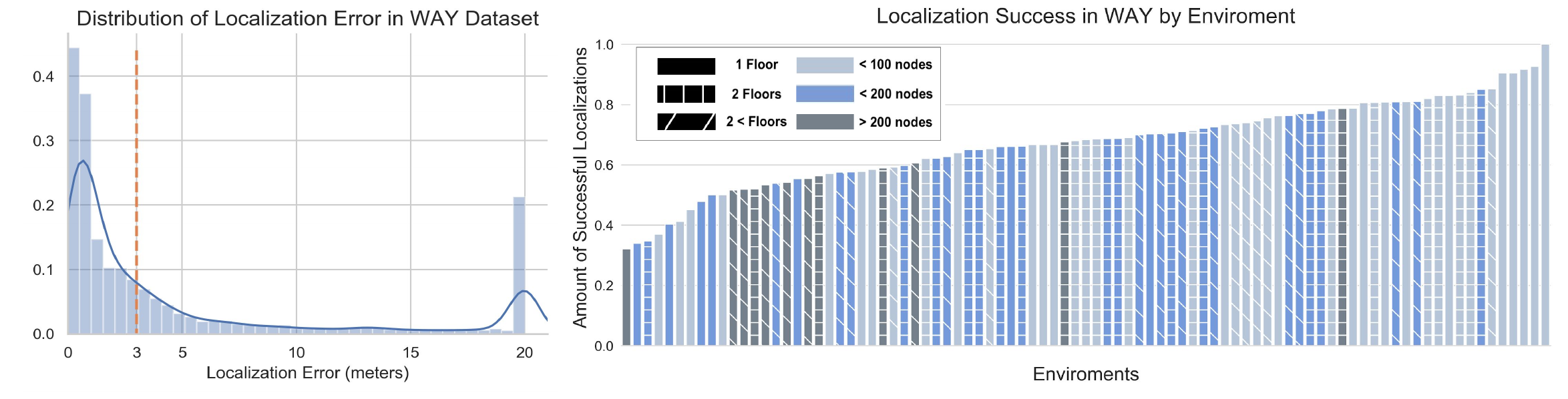}
\caption{Left: Distribution of human localization error in \acronym (20+ includes wrong floor predictions). Right: Human success rates (error $<$3m) by environment. Bar color indicates environment size (number of nodes) and pattern the number of floors. }
\label{fig:localizations}
\end{figure*}

\xhdr{Map Representation.}
The Locator is shown top-down views of Matterport textured meshes as environment maps. In order to increase the visibility of walls in the map (which may be mentioned by the Observer), we render views using perspective rather than orthographic projections (see left in \figref{fig:task_teaser}). We set the camera near and far clipping planes to render single floors such that multi-story buildings contain an image for each floor. 

\subsection{Collecting Human Localization Dialogs}
To provide a human-performance baseline and gather training data for agents, we collect human localization dialogs in these environments.  

\xhdr{Episodes.} 
We generate 2020 episodes across 87 environments by rejection sampling to avoid spatial redundancy. For each environment, we iteratively sample start locations, rejecting ones that are within 5m of already-sampled positions. Three environments were excluded due to their size (too large or small) or poor reconstruction quality. 

\xhdr{Data Collection.} 
We collect dialogs on Amazon Mechanical Turk (AMT) -- randomly pairing workers into \Obs or \Loc roles for each episode. The \Obs interface includes a first-person view of the environment and workers can pan/tilt the camera in the current position or click to navigate to adjacent panoramas. The \Loc interface shows the top-down map of the building, which can be zoomed and tilted to better display the walls. Views for each floor can be selected for multi-story environments. Both interfaces include a chat window where workers can send their message and end their dialog turn. The \Loc interface also includes the option to make their prediction by clicking a spot on the top-down map -- terminating the dialog. Note this option is only available after two rounds of dialog. Refer to the appendix for further details on the AMT interfaces. 

Before starting, workers were given written instructions and a walk-through video on how to perform their role. We restricted access to US workers with at least a 98\% success rate over 5,000 previous tasks. Further, we restrict workers from repeating tasks on the same building floor. In order to filter bad-actors, we monitored worker performance based on a running-average of localization error in meters and the number of times they disconnected from dialogs -- removing workers who exceeded a 10m threshold and discarding their data.

\xhdr{Dataset Splits.}
We follow the standard splits for the Matterport3D dataset \cite{chang2017matterport3d} -- dividing along environments. We construct four splits: train, val-seen, val-unseen, and test comprising 3,967/299/561/1,165 dialogs from 58/55/11/18 environments respectively. Val-seen contains new start locations for environments seen in train. Both val-unseen and test contain new environments. This allows us to assess generalization to new dialogs and to new environments separately in validation. Following best practices, the final locations of the observer for the test set will not be released but we will provide an evaluation server where predicted localizations can be uploaded for scoring.

\acronym includes dialogs in which the human \Loc failed to accurately localize the \Obs. In reviewing failed dialogs, we found human failures are often due to visual aliasing (e.g., across multiple floors), or are relatively close to the 3m threshold. We therefore expect that these dialogs still contain valid descriptions, especially when paired with the Observer’s true location during training. In experiments when removing failed dialogs from the train set, accuracy did not significantly change.

\subsection{Dataset Analysis}
\label{sec:analysis}

\xhdr{Data Collection and Human Performance.}
In total, 174 unique workers participated in our tasks. On average each episode took 4 minutes and the average localization error is 3.17 meters. Overall, 72.5\% of episodes where considered successful localizations at an error threshold of 3 meters. Each starting location has 3 annotations by separate randomly-paired Observer-Locator teams. In 40.9\% of start locations, all 3 teams succeeded, in 36.3\% 2, 18.5\% 1, and 4.3\% 0 teams succeeded. \figref{fig:localizations} left shows a histogram of localization errors.

\begin{figure*}[t]
\centering
\includegraphics[width=0.95\textwidth]{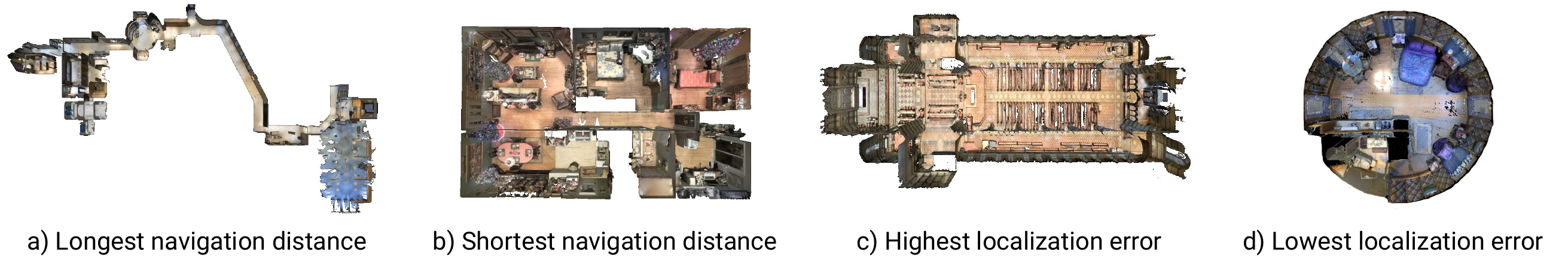}
\caption{Environments with the largest/smallest mean navigation distance (a, b) and mean localization error (c, d). Observers tend to navigate more in featureless areas, such as the long corridor in (a). Localization error is highest in buildings with many repeated indistinguishable features, such as the cathedral with rows of pews in (c).}
\label{fig:specific_buildings}
\end{figure*}

\xhdr{Why is it Difficult?}
Localization through dialog is a challenging task, even for humans. The teams success depends on the uniqueness of starting position, if and where the Observer chooses to navigate, and how discriminative the Locator's questions are. Additionally, people vary greatly in their ability to interpret maps, particularly when performing mental rotations and shifting perspective \cite{kozhevnikov2006perspective}, which are both skills required to solve this task. We also observe that individual environments play a significant role in human error -- as illustrated in \figref{fig:localizations} right, larger buildings and buildings with multiple floors tend to have larger localization errors, as do buildings with multiple similar looking rooms (e.g., multiple bedrooms with similar decorations or office spaces with multiple conference rooms). The buildings with the highest and lowest error are shown in \figref{fig:specific_buildings}.

\begin{figure*}[t]
\centering
\includegraphics[width=0.95\textwidth]{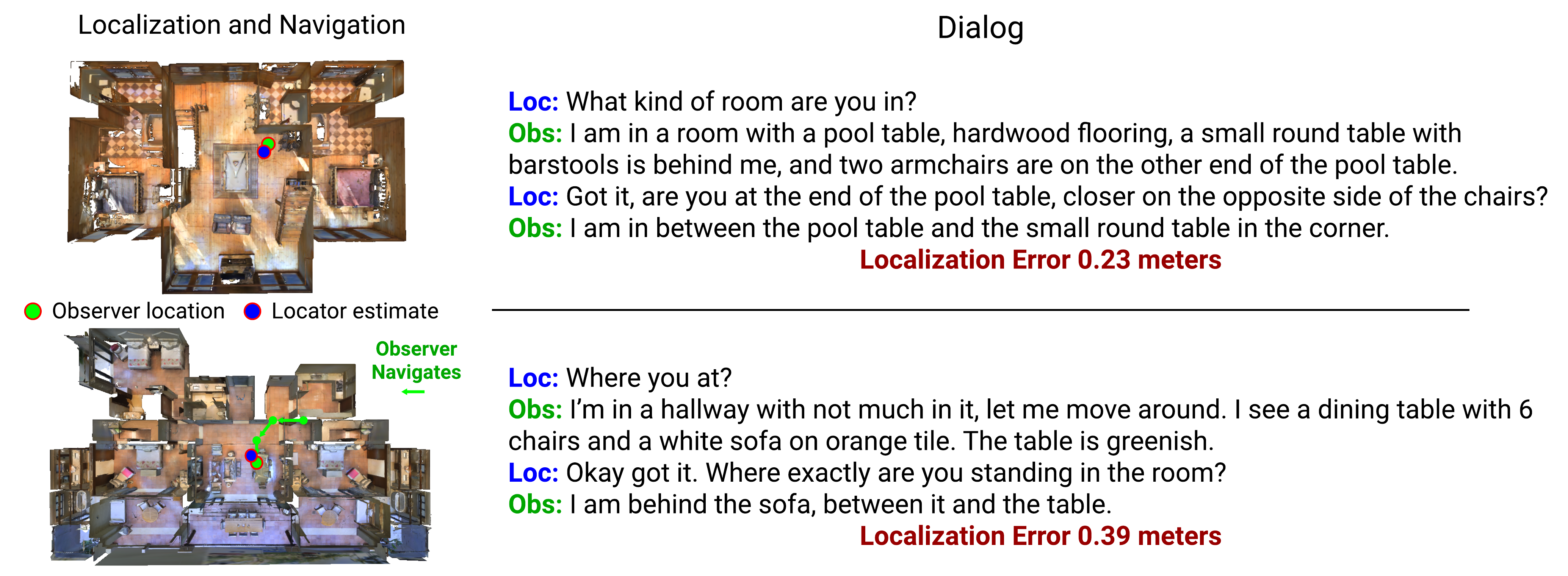}
\caption{Examples from the dataset illustrating the Observer's location on the top-down map vs. the Locator's estimate (left) and the associated dialog (right). In the bottom example the Locator navigates to find a more discriminative location, which is a common feature of the dataset. The Observer navigates in 63\% of episodes and the average navigation distance for these episodes is 3.4 steps (7.45 meters).}
\label{fig:dataset_teaser}
\end{figure*}

\xhdr{Characterizing \acronym Dialogs.} \figref{fig:dataset_teaser} shows two example dialogs from \acronym. These demonstrate a common trend -- the \Obs provides descriptions of their surroundings and then the \Loc asks clarifying questions to refine the position. More difficult episodes require multiple rounds to narrow down the correct location and the \Loc may ask the \Obs to move or look for landmarks. On average, dialogs contain 5 messages and 61 words. 

The \Obs writes longer messages on average (19 words) compared to the \Loc (9 words). This asymmetry follows from their respective roles. The \Obs has first-person access to high-fidelity visual inputs and must describe their surroundings, \myqoute{In a kitchen with a long semicircular black counter-top along one wall. There is a black kind of rectangular table and greenish tiled floor.}. Meanwhile, the \Loc sees a top-down view and uses messages to probe for discriminative details, \myqoute{Is it a round or rectangle table between the chairs?}, or to prompt movement towards easier to discriminate spaces, \myqoute{Can you go to another main space?}. 

As the \Loc has no information at the start of the episode, their first message is often a short prompt for the \Obs to describe their surroundings, further lowering the average word count. Conversely, the \Obs's reply is longer on average at 24 words. Both agent's have similar word counts for further messages as they refine the location. See the appendix for details on common utterances for both roles in the first two rounds of dialog.

\xhdr{Role of Navigation.} 
Often the localization task can be made easier by having the \Obs move to reduce uncertainty (see bottom example of \figref{fig:dataset_teaser}). This includes moving away from nondescript areas like hallways and moving to unambiguous locations. We observe at least one navigation step in 62.6\% of episodes and an average of 2.12 steps. Episodes containing navigation have a significantly lower average localization error (2.70m) compared to those that did not (3.98m). We also observe the intuitive trend that larger environments elicit more navigation. The distributions for start and end locations for the most and least navigated environments in the appendix.

\begin{figure*}[t]
\centering
\includegraphics[width=\textwidth]{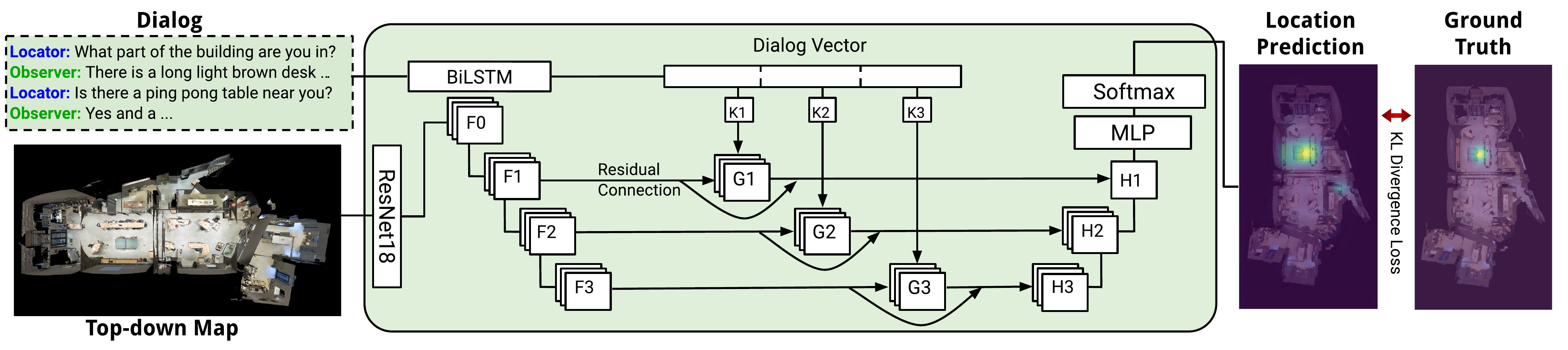}
\caption{The 3-layer LingUNet-Skip architecture used to model the Localization from Embodied Dialog task.}
\label{fig:lingunet}
\end{figure*}

\subsection{\dataset Tasks}
\label{sec:tasks}

We now formalize the LED, EVD and CL tasks to provide a clear orientation for future work.

\xhdr{Localization from Embodied Dialog.}
The LED task is the following -- given an episode comprised of a environment and human dialog -- $(e, L_0,O_0, \dots L_{T-1}, O_{T-1})$ -- predict the \Obs's final location ${p}_T$. This is a grounded natural language understanding task with pragmatic evaluations -- localization error and accuracy at a variable threshold which in this paper is set to 3 meters. This task does not require navigation or text generation; instead, it mirrors AI-augmented localization applications. An example would be a system that listens to emergency services calls and provides a real time estimate of the caller's indoor location to aid the operator. 

\xhdr{Embodied Visual Dialog.}
This task is to replace the Observer by an AI agent. 
Given a embodied first-person view of a 3D environment (see Observer view in Fig. \ref{fig:task_teaser}), and a partial history of dialog consisting of $k$ Locator and  $k-1$ Observer message pairs ($L_0$: \myqoute{describe your location.}, $O_0$: \myqoute{I'm in a kitchen with black counters.}, $L_1$ \dots): predict the Observer agent's next navigational action and natural language message to the Locator.
To evaluate the agent's navigation path, the error in the final location can be used along with path metrics such as nDTW \cite{ilharco2019general}. Generated text can be evaluated against human responses using existing text similarity metrics.

\xhdr{Cooperative Localization.}
In this task, both the Observer and the Locator are modeled agents. Modeling the Locator agent requires goal-oriented dialog generation and confidence estimation to determine when to end the task by predicting the location of the Observer. Observer and Locator agents can be trained and evaluated independently using strategies similar to the EVD task, or evaluated as a team using localization accuracy as in LED. 


\section{Modeling Localization From Embodied Dialog}
\label{sec:model}
While the \acronym dataset supports multiple tasks, we focus on Localization from Embodied Dialog as a first step. In LED, the goal is to predict the location of the \Obs given a dialog exchange. 

\setlength{\tabcolsep}{5pt}
\begin{table*}[t]
\begin{center}
\caption{Comparison of our model with baselines and human performance on the LED task. We report average localization error (LE) and accuracy at 3 and 5 meters (all $\pm$ standard error). {\footnotesize \textsuperscript{*} denotes oracle access to Matterport3D node locations.}}\vspace{-0.125in}
\footnotesize
\resizebox{\textwidth}{!}{
\label{table:results}
\begin{tabular}{l c ccc c ccc c csc}
\toprule
                &  & \multicolumn{3}{c}{val-seen}    & & \multicolumn{3}{c}{val-unseen}    & & \multicolumn{3}{c}{test}                                                                                                                                          \\\cmidrule(l{2pt}r{2pt}){3-5} \cmidrule(l{2pt}r{2pt}){7-9}  \cmidrule(l{2pt}r{2pt}){11-13}  
                
Method         & & \multicolumn{1}{c}{\scriptsize LE $\downarrow$}& \multicolumn{1}{c}{\scriptsize Acc@3m $\uparrow$}   & \multicolumn{1}{c}{\scriptsize Acc@5m $\uparrow$}   && \multicolumn{1}{c}{\scriptsize LE $\downarrow$}& \multicolumn{1}{c}{\scriptsize Acc@3m $\uparrow$}   & \multicolumn{1}{c}{\scriptsize Acc@5m $\uparrow$}   && \multicolumn{1}{c}{\scriptsize LE $\downarrow$}& \multicolumn{1}{c}{\scriptsize Acc@3m $\uparrow$}   & \multicolumn{1}{c}{\scriptsize Acc@5m $\uparrow$}   \\ \toprule
Human Locator&                   & 3.26\scriptsize{$\pm$0.71}    & 72.3\scriptsize{$\pm$3.0}              & 78.8\scriptsize{$\pm$3.0}     &         & 1.91\scriptsize{$\pm$0.32}              & 79.7\scriptsize{$\pm$3.0}              & 85.2\scriptsize{$\pm$1.7}      &    & 
3.16\scriptsize{$\pm$0.35}          & 
70.4\scriptsize{$\pm$1.4}            & 77.2\scriptsize{$\pm$1.3}          
\\\midrule
Random      &                    & 12.39\scriptsize{$\pm$0.31}   &  5.4\scriptsize{$\pm$0.9}              & 15.0\scriptsize{$\pm$1.3}         &     & 10.18\scriptsize{$\pm$0.16}          &  7.0\scriptsize{$\pm$0.7}              & 21.3\scriptsize{$\pm$1.1}     &     & 13.10\scriptsize{$\pm$0.17}    & 6.6\scriptsize{$\pm$0.5}           & 15.2\scriptsize{$\pm$0.7}          \\
Random Node\textsuperscript{*} & & 8.27\scriptsize{$\pm$0.44}     & 18.1\scriptsize{$\pm$2.2}              & 37.8\scriptsize{$\pm$2.7}      &        & 10.44\scriptsize{$\pm$0.31}           & 15.8\scriptsize{$\pm$1.1}              & 29.0\scriptsize{$\pm$1.4}    &      & 13.19\scriptsize{$\pm$0.32}        & 12.8\scriptsize{$\pm$0.7}          & 24.9\scriptsize{$\pm$0.9}          \\
Center     &                     & 6.13\scriptsize{$\pm$0.25}     & 23.1\scriptsize{$\pm$2.4}              & 46.5\scriptsize{$\pm$2.9}       &       & \textbf{4.90\scriptsize{$\pm$0.12}}              & 29.8\scriptsize{$\pm$1.9}              & 61.0\scriptsize{$\pm$2.1}     &     & \textbf{6.71\scriptsize{$\pm$0.14}}          & 22.6\scriptsize{$\pm$1.2}          & 42.3\scriptsize{$\pm$1.4}  \\
Heuristic     && 11.6\scriptsize{$\pm$0.49} & 12.5\scriptsize{$\pm$1.8} & 23.6\scriptsize{$\pm$2.4}   &&
10.10\scriptsize{$\pm$0.28}  & 10.5\scriptsize{$\pm$1.2} & 25.7\scriptsize{$\pm$1.8}     &&
13.45\scriptsize{$\pm$0.32}  & 9.1\scriptsize{$\pm$0.8}   & 18.4\scriptsize{$\pm$1.1} \\
No Language     && 7.17\scriptsize{$\pm$0.42}    & 26.1\scriptsize{$\pm$2.5} & 44.8\scriptsize{$\pm$2.9}       &&
5.72\scriptsize{$\pm$0.20}         & 32.1\scriptsize{$\pm$2.0}              & 58.1\scriptsize{$\pm$2.1}     &&
7.67\scriptsize{$\pm$0.18}          & 22.3\scriptsize{$\pm$1.2}          & 42.4\scriptsize{$\pm$1.4} \\ 
No Vision     && 11.36\scriptsize{$\pm$0.46}        &  9.4\scriptsize{$\pm$1.7}              & 18.4\scriptsize{$\pm$2.2}       &&
8.58\scriptsize{$\pm$0.20}         &  7.8\scriptsize{$\pm$1.1}              & 22.1\scriptsize{$\pm$1.8}     &&
11.62\scriptsize{$\pm$0.23}         & 7.7\scriptsize{$\pm$0.8}          & 18.3\scriptsize{$\pm$1.1}  

\\\midrule
LingUNet  &                      & \textbf{4.73\scriptsize{$\pm$0.32}}    & \textbf{53.5\scriptsize{$\pm$2.9}}              & \textbf{67.2\scriptsize{$\pm$2.7}}      &        & 5.01\scriptsize{$\pm$0.19}            & \textbf{45.6\scriptsize{$\pm$2.1}}              & \textbf{63.6\scriptsize{$\pm$2.0}}      &    & 7.32\scriptsize{$\pm$0.22}        & \textbf{32.7\scriptsize{$\pm$1.4}}          & \textbf{49.5\scriptsize{$\pm$1.5}}          \\ \bottomrule
\end{tabular}}
\end{center}
\end{table*}

\subsection{LED Model from Top-down Views}
We model localization as a language-conditioned pixel-to-pixel prediction task -- producing a probability distribution over positions in a top-down view of the environment. This choice mirrors the environment observations human Locators had during data collection, allowing straightforward comparison. However, future work need not be restricted to this choice and may leverage the panoramas or 3D reconstructions that Matterport3D provides. 

\xhdr{Dialog Representation.} \Loc and \Obs messages are tokenized using a standard toolkit \cite{loper2002nltk}. The dialog is represented as a single sequence with identical `start' and `stop' tokens surrounding each message, and then encoded using a single-layer bidirectional LSTM with a 300 dimension hidden state. Word embeddings are initialized using GloVe \cite{Pennington14glove:global} and finetuned end-to-end. 

\xhdr{Environment Representation.}
The visual input to our model is the environment map which we scale to 780$\times$455 pixels. We encode this map using a ResNet18 CNN \cite{he2015deep} pretrained on ImageNet \cite{ILSVRC15}, discarding the 3 final conv layers and final fully-connected layer in order to output a 98$\times$57 spatial map with feature dimension 128. Although the environment map is a top-down view which does not closely resemble ImageNet images, in initial experiments we found that using a pretrained and fixed CNN improved over training from scratch. 

\xhdr{Language-Conditioned Pixel-to-Pixel Model.} We adapt a language-conditioned pixel-to-pixel LingUNet \cite{misra2018mapping} to fuse the dialog and environment representations. We refer to the adapted architecture as LingUNet-Skip. As illustrated in \figref{fig:lingunet}, LingUNet is a convolutional encoder-decoder architecture. Additionally we introduce language-modulated skip-connections between corresponding convolution and deconvolution layers. 
Formally, the convolutional encoder produces feature maps $F_l = \text{Conv}(F_{l-1})$ beginning with the initial input $F_0$. Each feature map $F_l$ is transformed by a 1$\times$1 convolution with weights $K_l$ predicted from the dialog encoding, i.e.~$G_l = \text{Conv}_{K_l}(F_l)$. The language kernels $K_l$ are linear transforms from components of the dialog representation split along the feature dimension. Finally, the deconvolution layers combine these transformed skip-connections and the output of the previous layer, such that $H_{l} = \text{Deconv}([H_{l+1};(G_l+F_l)])$. There are three layers and the output of the final deconvolutional is processed by a MLP and a softmax to output a distribution over pixels.

\xhdr{Loss Function.} We train the model to minimize the KL-divergence between the predicted location distribution and the ground-truth location, which we smooth by applying a Gaussian with standard deviation of 3m (matching the success criteria). During inference, the pixel with highest probability is selected as the final predicted location. For multi-story environments, each floor is processed independently during training. During inference only the ground truth final floor is processed. This is done to maintain accurate euclidean distance measurements for localization error as euclidean distance is not meaningful when measuring across points on floors in multi-story environments. This schema is used for all baselines experiments except for human locators who select from all floors. 

\subsection{Experimental Setup}
\xhdr{Metrics.} We evaluate performance using localization error (LE) defined as the Euclidean distance in meters between the predicted Observer location $\hat{p}_T$ and the Observer's actual terminal location $p_T$: $\textnormal{LE} = ||p_T - \hat{p}_T||_2$. We also report a binary success metric that places a threshold $k$ on the localization error -- $\mathbbm{1}(\textnormal{LE} \leq k)$ -- for 3m and 5m. The 3m threshold allows for about one viewpoint of error since viewpoints are on average 2.25m apart. We use euclidean distance for LE because localization predictions are not constrained to the navigation graph. Matterport building meshes contain holes and other errors around windows, mirrors and glass walls, which can be problematic when computing geodesic distances for points off the navigation graph.

\xhdr{Training and Implementation Details.}
Our LingUNet-Skip model is implemented in PyTorch \cite{paszke2019pytorch}. Training the model involves optimizing around 16M parameters for 15--30 epochs, requiring $\sim$8 hours on a single GPU. We use the Adam optimizer \cite{kingma2014adam} with a batch size of $10$ and an initial learning rate of $0.001$ and apply Dropout \cite{srivastava2014dropout} in non-convolutional layers with $p=0.5$. We tune hyperparameters based on val-unseen performance and report the checkpoint with the highest val-unseen Acc@3m.  To reduce overfitting we apply color jitter, 180\degree~rotation, and random cropping by 5\% to the map during training.

\begin{figure*}[ht]
\centering
\includegraphics[ width=\textwidth]{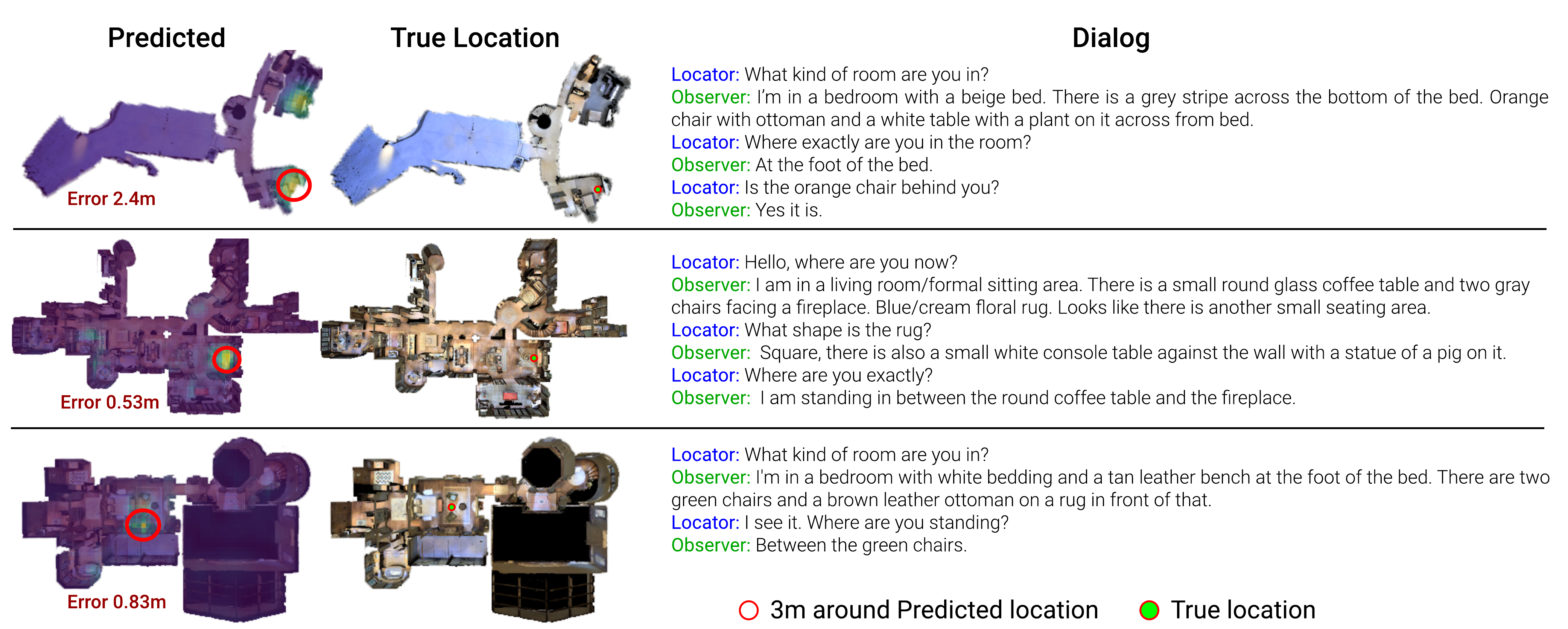}
\caption{Examples of the predicted distribution versus the true location over top down maps of environment floors for dialogs in val-unseen. The red circle on the left represents the three meter threshold around the predicted localization. The green dot on the middle image represents the true location. The localization error in meters of the predicted location is shown in red.}
\label{fig:examples_dist}
\end{figure*}

\xhdr{Baselines.} We consider a number of baselines and human performance to contextualize our results and analyze \acronym:
\begin{compactitem}[\hspace{3pt}--]
\item \textbf{Human Locator.} The average performance of AMT Locator workers as described in \secref{sec:dataset}.  
\item \textbf{Random.} Uniform random pixel selection.
\item \textbf{Center.} Always selects the center coordinate. 
\item \textbf{Random Node.} Uniformly samples from Matterport3D node locations. This uses oracle knowledge about the test environments. While not a fair comparison, we include this to show the structural prior of the navigation graph which reduces the space of candidate locations.
{\item \textbf{Heuristic Driven.}
For each dialog $D_t$ in the validation splits we find the most similar dialog $D_g$ in the training dataset based on BLEU score~\cite{papineni2002bleu}. From the top-down map associated with $D_g$, a 3m x 3m patch is taken around the ground truth Observer location. We predict the location for $D_t$ by convolving this patch with the top-down maps associated with $D_t$ and selecting the most similar patch (according to Structural Similarity). The results (below) are only slightly better than random.}
\end{compactitem}

\subsection{Results}
\tabref{table:results} shows the performance of our LingUNet-Skip model and relevant baselines on the val-seen, val-unseen, and test splits of the \acronym dataset.  

\xhdr{Human and No-learning Baselines.} Humans succeed 70.4\% of the time in test environments. Notably, val-unseen environments are easier for humans (79.7\%), see appendix for details.
The Random Node baseline outperforms the pixel-wise Random setting (Acc@3m and Acc@5m for all splits) and this gap quantifies the bias in nav-graph positions.
We find the Center baseline to be rather strong in terms of localization error, but not accuracy -- wherein it lags behind our learned model significantly (Acc@3m and Acc@5m for all splits).

\xhdr{LingUNet-Skip outperforms baselines.} Our LingUNet-Skip significantly outperforms the hand-crafted baselines in terms of accuracy at 3m -- improving the best baseline, Center, by an absolute 10\% (test) to 30\% (val-seen and val-unseen) across splits (a 45-130\% relative improvement). Despite this, it achieves higher localization error than the Center model for val-unseen and test. This is a consequence of our model occasionally being quite wrong despite its overall stronger localization performance. There remains a significant gap between our model and human performance -- especially on novel environments (70.4\% vs 32.7\% on test).

\begin{table}[t]
\footnotesize
\begin{center}
\caption{Modality, modeling, and dialog ablations for our LingUNet-Skip model on the validation splits of \acronym.}\vspace{-0.125in}
\label{table:lang_ablations}
\resizebox{\columnwidth}{!}{
\begin{tabular}{lcccs}
\toprule
                    & \multicolumn{2}{c}{val-seen}                                & \multicolumn{2}{c}{val-unseen}                               \\ \cmidrule(l{2pt}r{2pt}){2-3} \cmidrule(l{2pt}r{2pt}){4-5}
            & \multicolumn{1}{c}{\scriptsize LE $\downarrow$} & \multicolumn{1}{c}{\scriptsize Acc@3m $\uparrow$}  & \multicolumn{1}{c}{\scriptsize LE $\downarrow$} & \multicolumn{1}{c}{\scriptsize Acc@3m $\uparrow$} \\ \toprule
Full LingUNet-Skip Model     &\textbf{4.73\scriptsize{$\pm$0.32}} & \textbf{53.5\scriptsize{$\pm$2.9}} & 5.01\scriptsize{$\pm$0.19}         & 45.6\scriptsize{$\pm$2.1}          \\
~~~w/o Data Aug.        &  
5.98\scriptsize{$\pm$0.35}    &
41.1\scriptsize{$\pm$2.0}              & 5.44\scriptsize{$\pm$0.18}  & 
35.7\scriptsize{$\pm$2.1}                                \\
~~~w/o Residual         &  5.26\scriptsize{$\pm$0.33}        & 47.5\scriptsize{$\pm$2.9}          & 4.74\scriptsize{$\pm$0.17}         & 43.1\scriptsize{$\pm$2.1}          \\\midrule
~~~No Dialog            &  7.17\scriptsize{$\pm$0.42}        & 26.1\scriptsize{$\pm$2.5}          & 5.72\scriptsize{$\pm$0.20}         & 32.1\scriptsize{$\pm$2.0}          \\
~~~First-half Dialog    &  5.06\scriptsize{$\pm$0.33}        & 50.5\scriptsize{$\pm$2.8}          & \textbf{4.71\scriptsize{$\pm$0.18}}& \textbf{46.2\scriptsize{$\pm$2.1}} \\
~~~Second-half Dialog   &  5.29\scriptsize{$\pm$0.28}        & 41.8\scriptsize{$\pm$2.8}          & 5.06\scriptsize{$\pm$0.17}         & 38.7\scriptsize{$\pm$2.1}          \\
~~~Observer-only        &  5.73\scriptsize{$\pm$0.36}        & 45.2\scriptsize{$\pm$2.9}          & 4.77\scriptsize{$\pm$0.17}         & 44.9\scriptsize{$\pm$2.1}          \\
~~~Locator-only         &  6.39\scriptsize{$\pm$0.37}        & 30.4\scriptsize{$\pm$2.7}          & 5.63\scriptsize{$\pm$0.19}         & 33.3\scriptsize{$\pm$2.0}          \\
~~~Shuffled Rounds      &  5.32\scriptsize{$\pm$0.32}        & 42.8\scriptsize{$\pm$2.8}          & 4.67\scriptsize{$\pm$0.18}         & 44.9\scriptsize{$\pm$2.1}          \\ \bottomrule
\end{tabular}}
\end{center}
\end{table}

\subsection{Ablations and Analysis}
Tab. \ref{table:lang_ablations} reports detailed ablations of our LingUNet-Skip model. Following standard practice, we report performance on val-seen and val-unseen.

\xhdr{Navigation Nodes Prior} 
We do not observe significant differences between val-seen (train environments) and val-unseen (new environments), which suggests the model is not memorizing the node locations. Even if the model did, learning this distribution would likely amount to free-space prediction which is a useful prior in localization.

\xhdr{Input Modality Ablations.} No Vision explores the extent that linguistic priors can be exploited by LingUNet-Skip, while No Dialog does the same for visual priors. No Dialog beats the Center baseline (32.1\% vs. 29.8\% val-unseen Acc@3m) indicating that it has learned a visual centrality prior that is stronger than the center coordinate. This makes sense because some visual regions like nondescript hallways are less likely to contain terminal Observer locations. Both No Vision and No Dialog perform much worse than our full model (7.8\% and 32.1\% val-unseen Acc@3m vs. 45.6\%), suggesting that the task is strongly multimodal. 

\xhdr{Dialog Halves.} First-half Dialog uses only the first half of dialog pairs, while Second-half Dialog uses just the second half. Together, these examine whether the start or the end of a dialog is more salient to our model. We find that First-half Dialog performs marginally better than using the full dialog (46.2\% vs 45.6\% val-unseen Acc@3m) which we suspect is due to our model's failure to generalize second half dialog to unseen environments and problems handling long sequences. Further intuition for these results is that the first-half of the dialog contains coarser grained descriptions and discriminative statements (“I am in a kitchen”). The second-half of the dialog contains more fine grained descriptions (relative to individual referents in a room). Without the initial coarse localization, the second-half dialog is ungrounded and references to initial statements are not understood, therefore leading to poor performance.

\xhdr{Observer dialog is more influential.} Observer-only ablates Locator dialog and Locator-only ablates Observer dialog. We find that Observer-only significantly outperforms Locator-only (44.9\% vs. 33.3\% val-unseen Acc@3m). This is an intuitive result as Locators in the WAY dataset commonly query the Observer for new information. We note that Observers were guided by the Locators in the collection process (e.g. \myqoute{What room are you in?}), and that ablating the Locator dialog does not remove this causal influence.

\xhdr{Shuffling Dialog Rounds.} Shuffle Rounds considers the importance of the order of Locator-Observer dialog pairs by shuffling the rounds. Shuffling the rounds causes our LingUNet-Skip to drop just an absolute 0.7\% val-unseen Acc@3m (2\% relative).

\xhdr{Model Ablations.} Finally, we ablate two model-related choices. Without data augmentation (w/o Data Aug.), our model drops 9.9\% val-unseen Acc@3m (22\% relative). Without the additional residual connection (w/o Residual), our model drops 2.5\% val-unseen Acc@3m (5\% relative).

\section{Conclusion and Future Work}
\label{sec:conclusion}
In summary, we propose a new set of embodied localization tasks: Localization from Embodied  Dialog - LED (localizing the Observer from dialog history), Embodied Visual Dialog - EVD (modeling the Observer), and Cooperative Localization - CL (modeling both agents). 
To support these tasks we introduce \dataset a dataset containing $\sim$6k human dialogs from a cooperative localization scenario in a 3D environment. \acronym is the first dataset to present extensive human dialog for an embodied localization task. On the LED task we show that a LingUNet-Skip model improves over simple baselines and model ablations but without taking full advantage of the second half of the dialog. Since \acronym encapsulates multiple embodied localization tasks, there remains much to be explored.

\section*{Acknowledgments} Partial funding for this work was provided by NIH award R01MH114999.

\bibliography{anthology, emnlp2020}
\bibliographystyle{acl_natbib}
\clearpage
\section{Appendix}
\label{sec:supp}
\xhdr{Val-Unseen has higher accuracy than other splits.}
Human’s localization Acc@3m is 79.4\% for val-unseen which is higher than all other splits such as test which as a Acc@3m of 70.4\%. Following standard practice, the splits followed \cite{chang2017matterport3d}. The val-unseen split is notably smaller than the rest of the splits and through qualitative analysis, we found that the environments in the val-unseen split \cite{chang2017matterport3d} are generally smaller and have discriminative features which we attribute to the split having a high localization performance. Our LingUNet-Skip model has lower performance on test than on val-unseen which we reason is be to the nature of the environments in the splits. Additionally the LingUNet-Skip model has lower performance on test than on val-seen which is expected because test environments are unseen environments and val-seen environments are contained in the training set.

\begin{figure*}[ht]
\centering
\includegraphics[width=.8\textwidth]{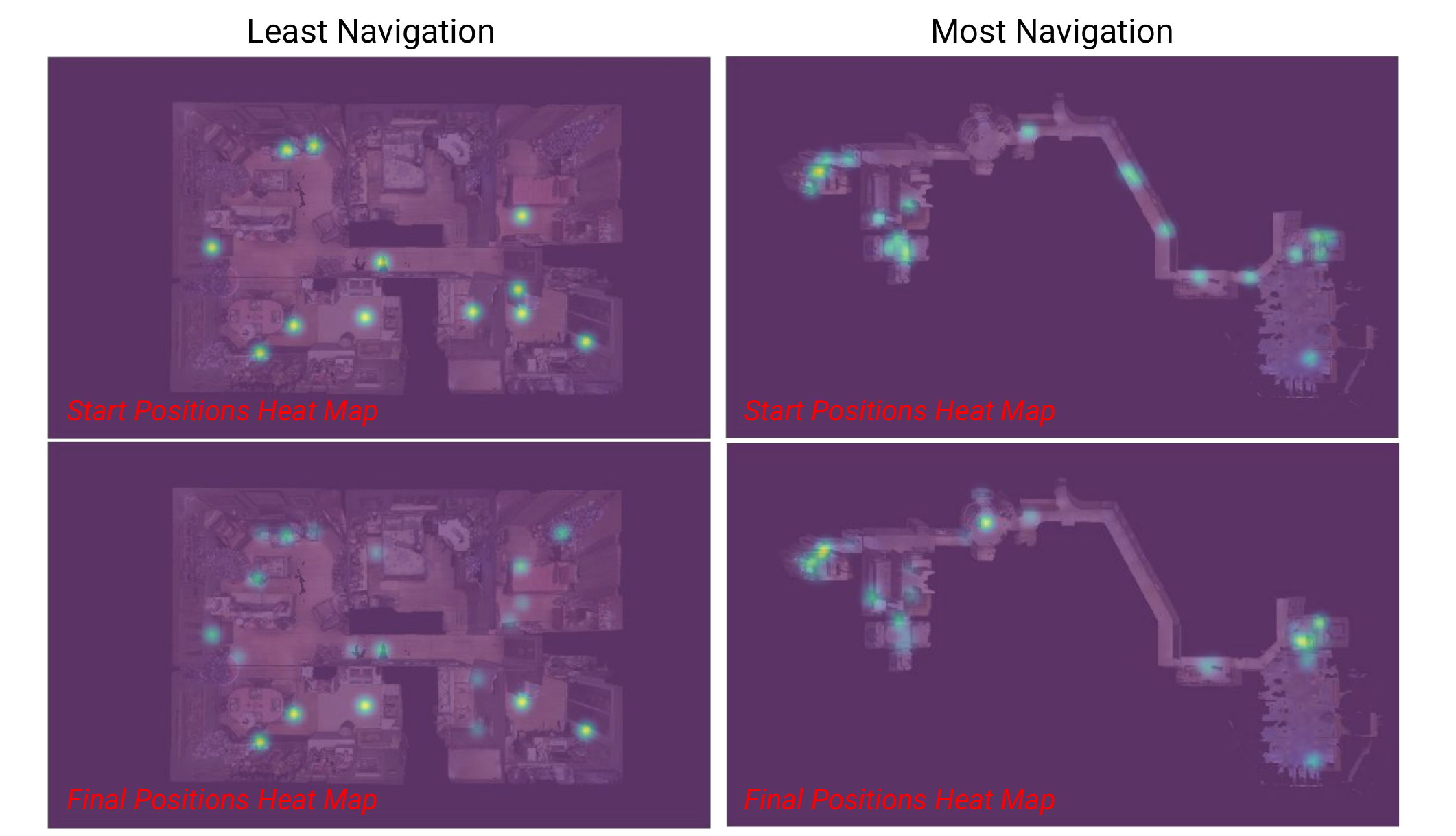}
\caption{Shows the distribution of the starting and ending locations of the Observer for two environments in the WAY dataset. On the left is the environment that had annotations which the least amount of navigation. On the right is the environment that had annotations with the most amount of navigation. }
\label{fig:start_end_positions}
\end{figure*}

\xhdr{Navigation differs between environments.}
As previously discussed, different environments in the WAY dataset have varying levels of navigation. This is likely attributed to a few factors such as size of the building and discriminative features of the building such as decorations. Additionally we see features like long hallways frequently lead to long navigational paths. The variances in navigation between environments is further illustrated in Fig. \ref{fig:start_end_positions}. While the distribution between the starting and final positions barely changes for the environment on the left, we see significant change in the environment on the right. Most noticeably we see that there are no final positions in the long corridor of the right environment despite it containing several start locations.

\begin{figure*}
\centering
\includegraphics[width=0.95\textwidth]{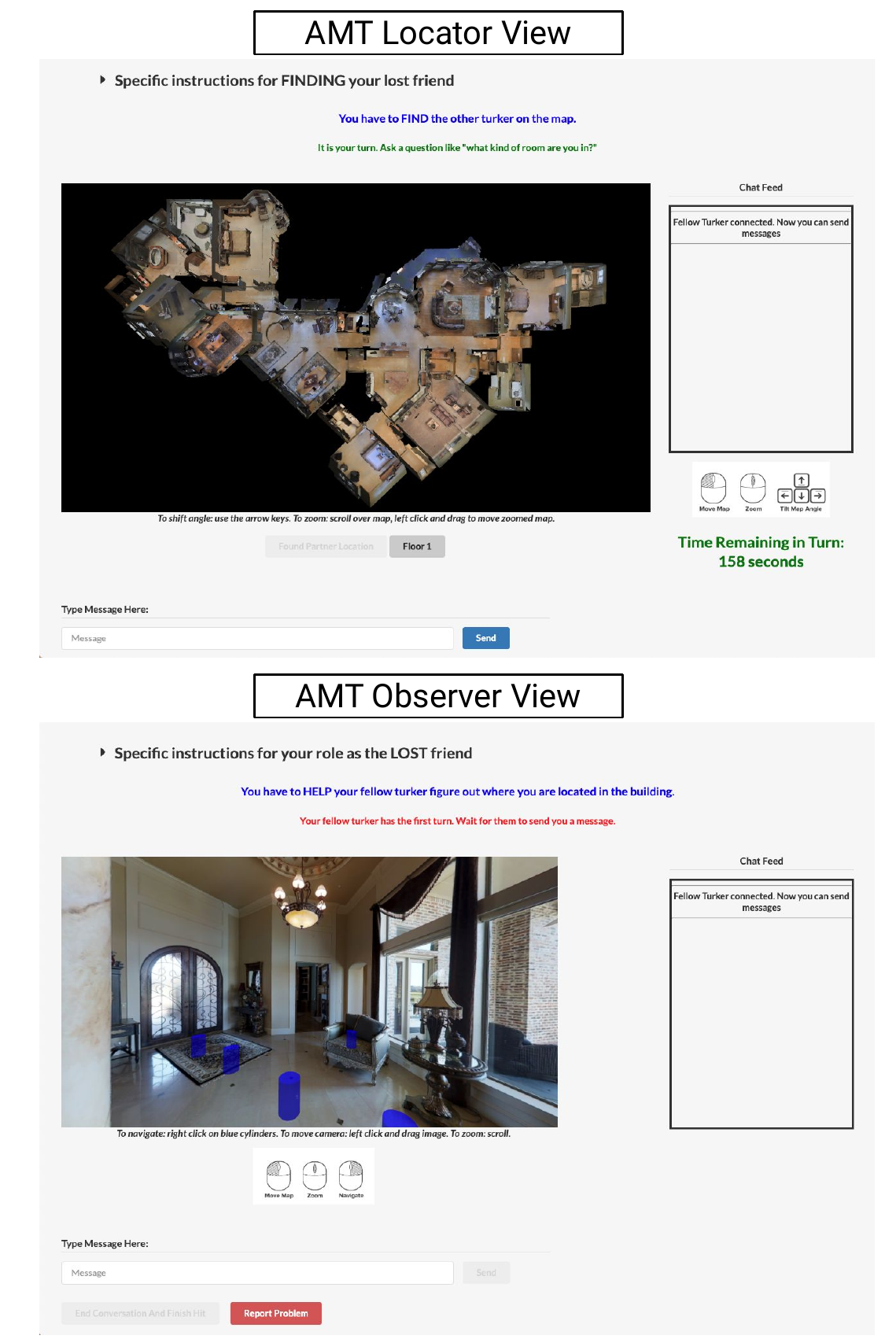}
\caption{The dataset collection interface for WAY. These are the interfaces that the Observer and Locator workers used on Amazon Mechanical Turk.}
\label{fig:amt_views}
\end{figure*}

\xhdr{Data Collection Interface.}
Fig. \ref{fig:amt_views} shows the data collection interface for the Observer and Locator human annotators. The annotator team was able to chat with each other via a message box that also displayed the chat history. The Locator had a top down map of the environment and had buttons to switch between floors. The Observer was given a first person view of the environment and could navigate the environment by clicking on the blue cylinders shown in Fig. \ref{fig:amt_views}

\begin{figure*}
\centering
\includegraphics[width=1\textwidth]{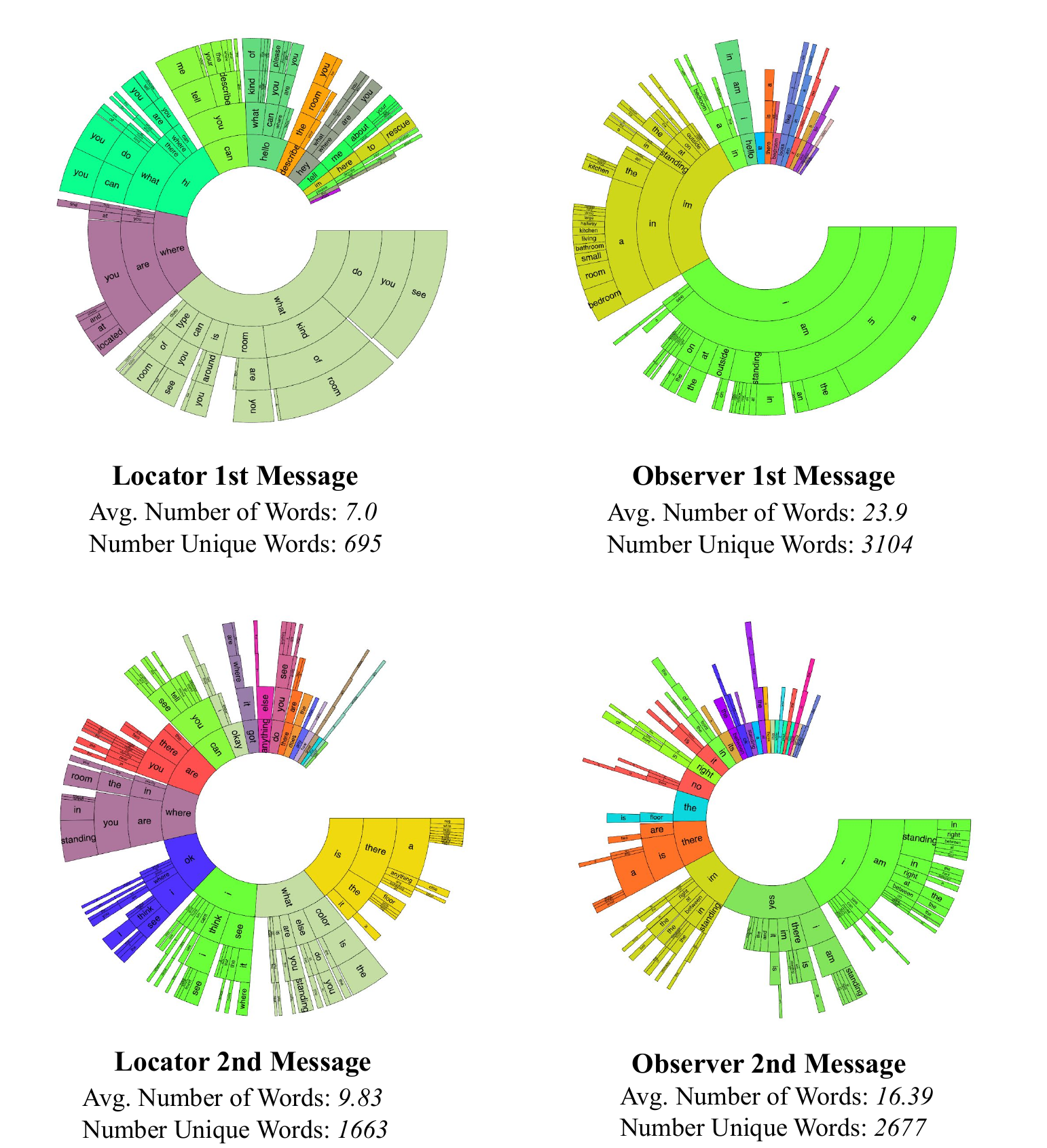}
\caption{Distributions of the first four words for each of the first four messages of the dialogs in the WAY dataset separated by message number and role type. The ordering of the words starts in the center and radiates outwards. The arc length is proportional to the number of messages containing the word. The white areas are words that had too low of a count to illustrate.}
\label{fig:sunbursts}
\end{figure*}

\xhdr{Closer Look at Dialog.}
Fig. \ref{fig:sunbursts} further breaks this down by looking at the average length of specific messages of the two agents. 
The Locator's first message is short in comparison to the average number of words per message of the agent. This is expected as this message is always some variation of getting the Observer to describe their location and it follows that the message has a low number of unique words. The Observer's first message is by far their longest, at 23.9 words, which is logical since in this message the Observer is trying to give the most unique description possible with no constraint on length. The distributions become more uniform in the 2nd messages from both the Locator and Observer. 
While the first message of the observer has a large number of unique words the distribution is not uniform over the words leading to the conclusion that the message has an common structure to it but that the underlying content is still discriminative for modeling the location. The word distribution of messages further down in the dialogue sequence are largely conditioned on the previous message from the other agent, which means that accurately encoding the dialogue history is important for accurate location estimation.

\begin{figure*}[ht]
\centering
\includegraphics[width=1\textwidth]{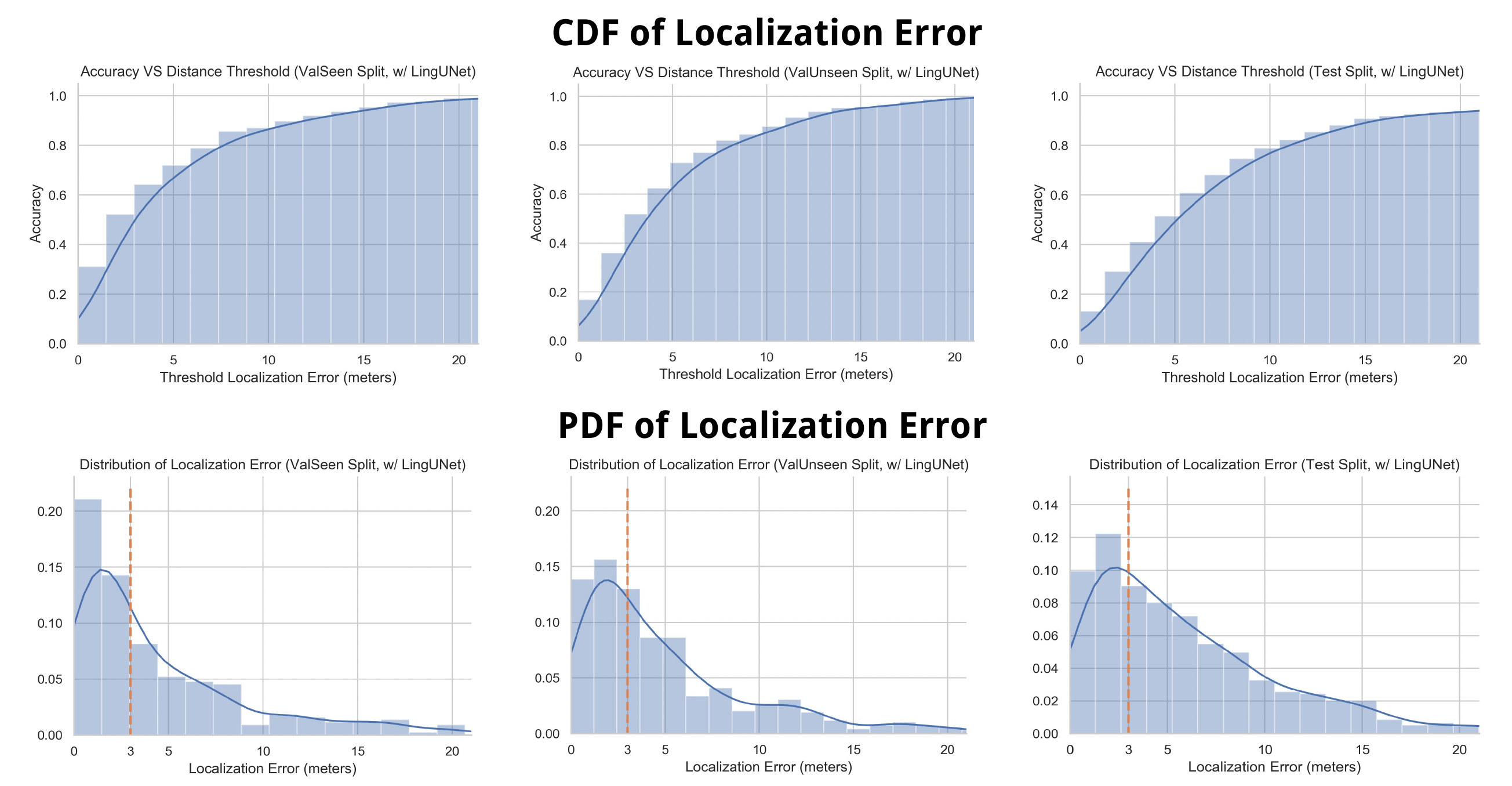}
\caption{The top row is the cdf of localization errors on the val and test splits using the LingUNet-Skip model. These graphs can also be interpreted as the accuracy vs threshold of error which defines success. The bottom row is the probability distribution of localization errors from the LingUNet-Skip model across the val and test splits.}
\label{fig:dist_error}
\end{figure*}

\xhdr{Distribution of Localization Error.} In order to better understand the distribution of the LingUNet-Skip model's predictions we visualize the distributions in Fig. \ref{fig:dist_error}.

\begin{figure*}[ht]
\centering
\includegraphics[ width=1\textwidth]{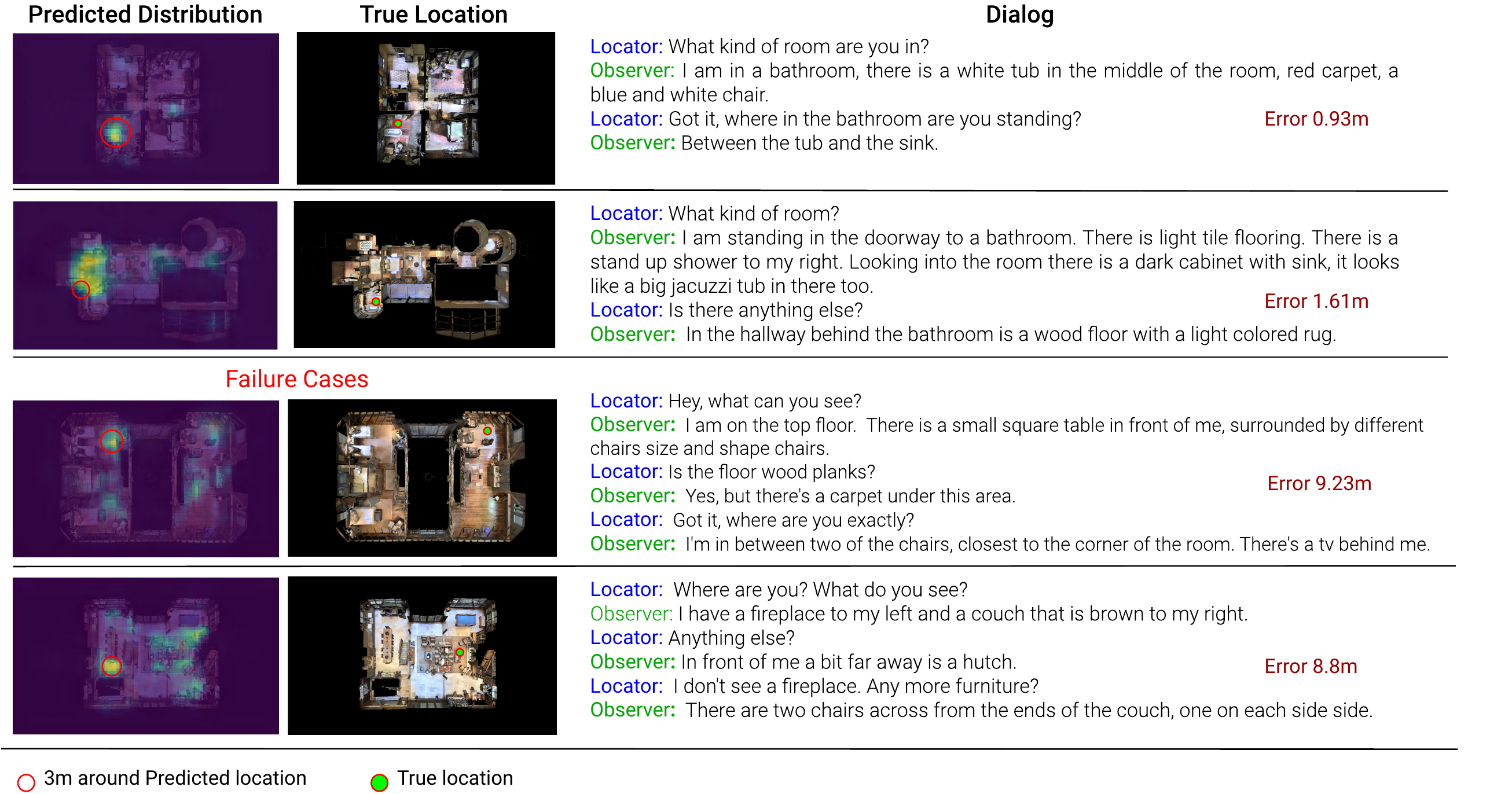}
\caption{Examples of the predicted distribution versus the true location over top down map of a floor of an environment for a given dialog in val-unseen. On the left the red circle represents the three meter threshold around the predicted localization. On the middle image the green dot represents the true location. The localization error in meters of the predicted location shown in red in the dialog box.}
\label{fig:supp_model_examples}
\end{figure*}

\xhdr{Success and Failure Examples.} To qualitatively evaluate our model we visualize the predicted distributions, the true location over the top down map and the dialog in Fig. \ref{fig:supp_model_examples}. We also show two failure cases in which the model predicts the wrong location.

\begin{figure*}[ht]
\centering
\includegraphics[ width=1\textwidth]{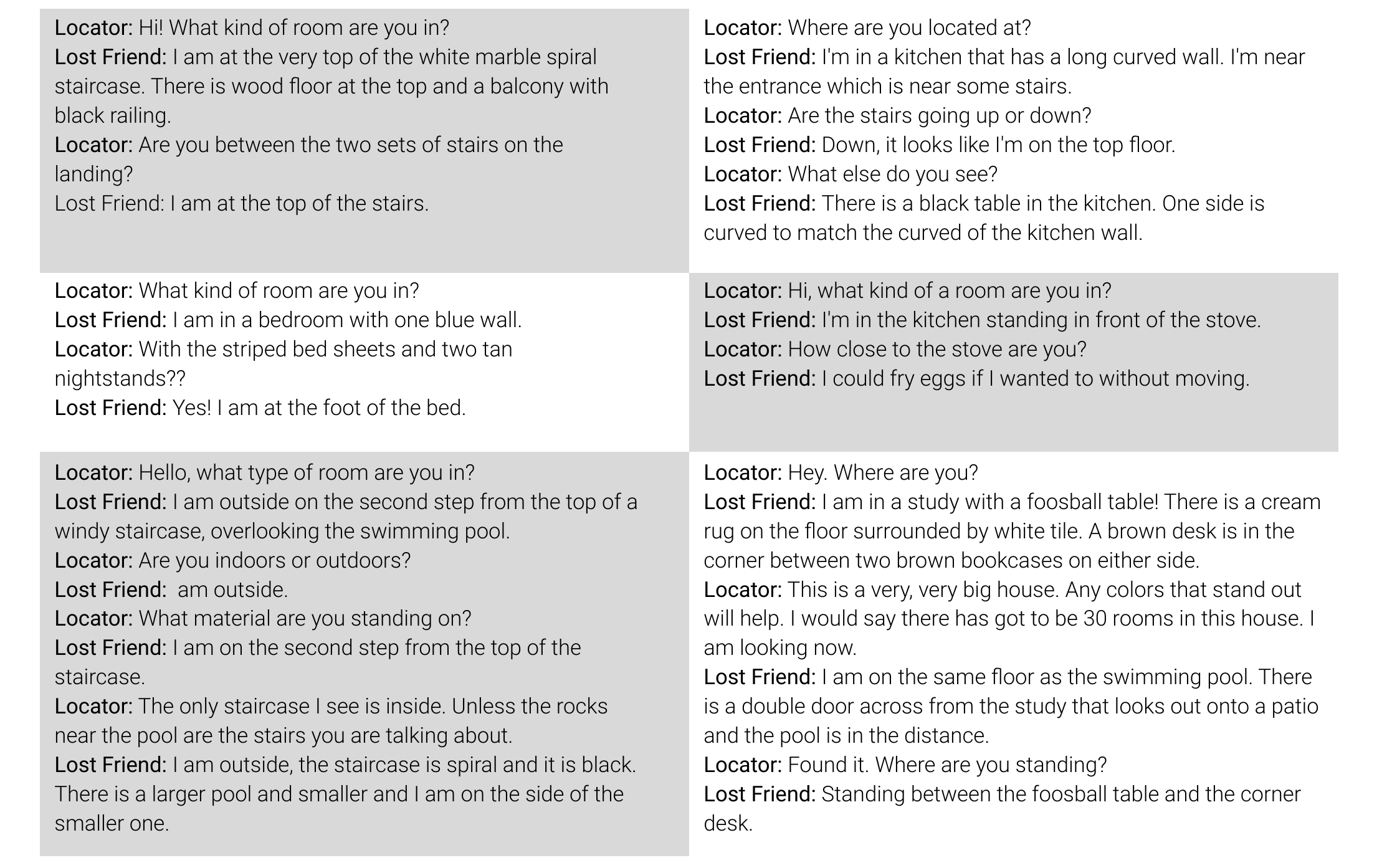}
\caption{Examples of dialog in the WAY dataset.}
\label{fig:supp_dialog_examples}
\end{figure*}

\end{document}